\documentclass[letterpaper, 10 pt, conference]{ieeeconf}  % Comment this line out if you need a4paper

\IEEEoverridecommandlockouts                              % This command is only needed if 
\usepackage[dvipsnames]{xcolor}
\newcommand{\changeF}[1]{\textcolor{black}{#1}}

\usepackage{amsmath}
\usepackage[pdftex]{graphicx}
\usepackage{tabularx}
\usepackage{multirow}
\usepackage{wrapfig}
\usepackage{booktabs}
\usepackage{amsmath,amssymb}
\usepackage{caption}
\usepackage{subcaption}

\usepackage{flushend}
\usepackage[export]{adjustbox}

\title{\LARGE \bf
Neural Transformers for Intraductal Papillary Mucosal Neoplasms (IPMN) Classification in MRI images
}

\author{F. Proietto Salanitri$^{1}$, G. Bellitto$^{1},$  S. Palazzo$^{1}$, I. Irmakci$^{2}$, M. Wallace$^{4}$, C. Bolan$^{4}$, M. Engels$^{4}$, \\ S. Hoogenboom$^{4}$, M. Aldinucci$^{3}$, U. Bagci$^{2}$, D. Giordano $^{1}$ and C. Spampinato$^{1}$% <-this % stops a space
\thanks{$^{1}$PeRCeiVe Lab, Department of Electrical, Electronic and Computer Engineering, University of Catania, Italy. 
        {\tt\small perceive@dieei.unict.it}}%
\thanks{$^{2}$Department of Radiology and BME, Northwestern University, Chicago, IL, USA}%
\thanks{$^{3}$ Computer Science Department , University of Torino, Torino, Italy}%
\thanks{$^{4}$ Mayo Clinic, Jacksonville, FL, USA}%
}

\begin{document}

\maketitle
\thispagestyle{empty}
\pagestyle{empty}

%%%%%%%%%%%%%%%%%%%%%%%%%%%%%%%%%%%%%%%%%%%%%%%%%%%%%%%%%%%%%%%%%%%%
\begin{abstract}
Early detection of precancerous cysts or neoplasms, i.e., Intraductal Papillary Mucosal Neoplasms (IPMN), in pancreas is a challenging and complex task, and it may lead to a more favourable outcome. 
Once detected, grading IPMNs accurately is also necessary, since low-risk IPMNs can be under surveillance program, while high-risk IPMNs have to be surgically resected before they turn into cancer.
Current standards (Fukuoka and others) for IPMN classification show significant intra- and inter-operator variability, beside being error-prone, making a proper diagnosis unreliable.
The established progress in artificial intelligence, through the deep learning paradigm, may provide a key tool for an effective support to medical decision for pancreatic cancer. In this work, we follow this trend, by proposing a novel AI-based IPMN classifier that leverages the recent success of transformer networks in generalizing across a wide variety of tasks, including vision ones. We specifically show that our transformer-based model exploits pre-training better than standard convolutional neural networks, thus supporting the sought architectural universalism of transformers in vision, including the medical image domain and it allows for a better interpretation of the obtained results.  
%This electronic document is a ÒliveÓ template. The various components of your paper [title, text, heads, etc.] are already defined on the style sheet, as illustrated by the portions given in this document.
%\newline

%\indent \textit{Clinical relevance}— This is a brief additional statement on why a this might be of interest to practicing clinicians. Example: This establishes the anesthetic efficacy of 10\% intraosseous injections with epinephrine to positively influence cardiovascular function.
\end{abstract}

%%%%%%%%%%%%%%%%%%%%%%%%%%%%%%%%%%%%%%%%%%%%%%%%%%%%%%%%%%%%%%%%%%%%%%%%%%%%%%%%
\section{INTRODUCTION}

\label{sec:intro}
Pancreatic cancer, also known as pancreatic ductal adenocarcinoma (PDAC), is a growing public health issue around the world. In the United States in 2021, an estimated 60,430 new cases of pancreatic cancer will be diagnosed, with 48,220 people dying from the disease \cite{society2021cancer}.
Pre-cancerous cysts or neoplasms in the pancreatic ducts are known as Intraductal Papillary Mucosal Neoplasms (IPMN) and can develop anywhere in the pancreas' ductal zone. Grading the severity of IPMNs is an important diagnosis step: most IPMNs are low-grade, and should be monitored over time; high-grade IPMNs, however, may turn into invasive cancer if left untreated. In these cases, surgery is the first choice to prevent them from expanding into malignant pancreatic tumors.
Therefore, there is an unmet for early detection techniques of IPMNs, in order to identify which IPMNs may lead to cancer.
Automated image analysis in radiology imaging plays a key role in diagnosis, treatment and intervention of pancreas diseases; thus there is a strong potential for machine learning tools to  support IPMN grade prediction that can serve better than the current radiographic standards.
The most popular imaging modalities for the pancreas are computed tomography (CT) and magnetic resonance imaging (MRI).  \\
In the last few years, transformer architectures~\cite{vaswani2017attention,dosovitskiy2020image} have raised as a valid alternative to standard convolutional networks on a variety of different tasks. 
More specifically, transformers enable learning arbitrary functions and consists of two main operation blocks: first, an attention-based block for modeling inter-element relations; second, a multi-layer perceptron (MLP) modeling relations intra-element. A sequence of attention and MLP blocks intertwined with residual connections and normalization has showed to allow for generalization over multiple tasks.  
Following this trend, herein we propose an automated IPMN classifier based on transformer architecture. We, in particular, show how transformers generalize better than standard and state-of-the-art CNNs (namely, DenseNet, AlexNet, etc.) also for extremely complex tasks, as IPMN classification, while providing similar accuracy to the state of the art IPMN classification study with deep learning~\cite{lalonde2019inn}. \\
The major contributions of this study are the following:
\begin{enumerate}
    \item Our work on IPMN classification is an important application contribution, which is not widely done due to the difficulty nature of the problem, and hence there is a very limited published research on this task using MRI data with deep learning.  
    Our  method  can provide a significant state-of-the-art baseline to be compared with for further MRI pancreas research just before critical surgery decision or surveillance.
    \item Our study contributes to the recent AI research in the strive to demonstrate architectural universalism of Transformers that can be used in a wide variety of tasks using little inductive bias, beside validating their better interpretability than CNN counterparts. To our best of knowledge, transformers have never been tried on high-risk medical diagnoses tasks before, particularly for pancreas imaging research.    
\end{enumerate}

\section{RELATED WORK}
\label{sec:related}

While significant progress has been made for automated approaches to segment the pancreas and its cysts~\cite{zhou2017deep}, the use of advanced machine learning algorithms to perform fully automatic risk-stratification of IPMNs is still limited. \\
Some recent works, employing machine learning techniques for predicting the risk of malignancy in IPMN, have used  endoscopic ultrasound (EUS) images \cite{kuwahara2019usefulness,gorris2021artificial} yielding high accuracy of 94.0\%, outperforming both human diagnosis (56\%) and conventional guidelines (40–68\%). 
CT imaging has been also adopted for investigating IPMN as in~\cite{hanania2016quantitative,gazit2017quantification} where low-level imaging features, such as texture, strength, and shape, have extracted from segmented cysts or pancreas for IPMN classification. 
Recently, deep learning methods based on standard convolutional neural networks have been proposed to diagnose IPMN from MRI scans~\cite{hussein2018deep,corral2019deep,lalonde2019inn}. 
Sarfaraz et. al.~\cite{hussein2018deep} proposed an architecture for automated IPMN classification based on feature extraction with canonical correlation using a pre-trained 3D CNN, while~\cite{corral2019deep} propose a novel CNN for recognizing high grade dysplasia or cancer on MR-images, yielding promising results. 
Finally, Rodney et al.~\cite{lalonde2019inn} constructed two novel ''inflated'' CNN architectures, InceptINN and DenseINN, for the task of diagnosing IPMN from multisequence (T1 and T2) MRI obtaining an accuracy of about 73\% in grading IPNM into three classes (no risk, low and high-risk). %by using the same data we employ in this work.\\ 
In this work, we employ transformers that are specific neural architectures originally proposed for machine translation tasks~\cite{vaswani2017attention}. Transformer-based models in NLP are generally pre-trained on large corpora and then fine-tuned for the task at hand~\cite{devlin2018bert,radford}.
Their increasing interest to vision tasks starts with Vision Transformers~\cite{dosovitskiy2020image} and Detection Transformer~\cite{carion2020end}. \changeF{Recently, several methods have explored transformer-based architectures for medical image  analysis mainly for  segmentation tasks~\cite{chen2021transunet,xie2021cotr,hatamizadeh2022unetr}. However, these method employ an hybrid architecture combining both convolutions and transformers. } 
Our approach builds upon pure vision transformers and employs a strategy similar to that one employed in NLP (as in~\cite{devlin2018bert,radford}), i.e., pre-training transformers on natural images and then fine-tuning them to MRI IPNM images. 
Experimental results show that our pre-trained transformers perform significantly better than state-of-the-art CNN classifiers.

\section{METHOD}
\label{sec:method}

\begin{figure}[htb]
  \centering
  \centerline{\includegraphics[width=8.5cm]{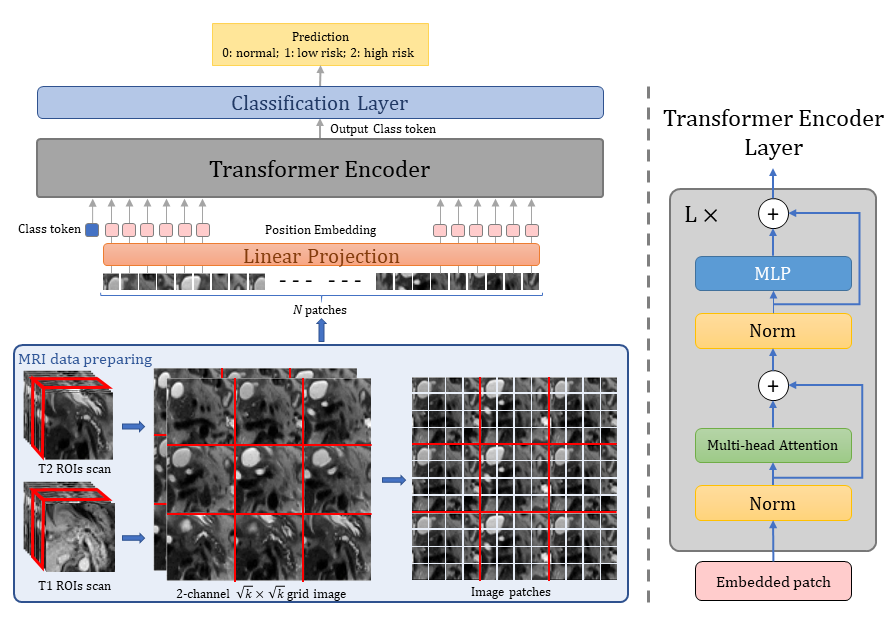}}
\caption{The proposed transformer-based architecture. T1 and T2 slices are concatenated along the channel dimension and sequences of 9 consecutive slices are arranged in a 3$\times$3 grid. Patches are then extracted from the resulting image, and are used as input to the transformer architecture. After encoding the patch set through transformer layers (consisting of a cascade of multihead attention block and MLP layers), a special \emph{classification token} encodes global image representation, and is used for final classification into three IPMN classes: \textit{normal}, \textit{low risk} and \textit{high risk}.
\label{fig:net}
}
\end{figure}

In our study, we follow the recently emerging approach of \textit{Transformers}~\cite{vaswani2017attention} for vision tasks. In particular, we use the ViT~\cite{dosovitskiy2020image} setting, in which the encoder of the original transformer model is used on a sequence of image ``patches''. However, since~\cite{dosovitskiy2020image} is trained on natural images, it is necessary to adapt the input representation to be able to process MRI scans, which are instead composed by an aggregation of multiple slices, providing anatomical volumetric information.

Fig.~\ref{fig:net} describes the proposed procedure in detail. We use T1- and T2-weighted MRI scans of the same patients in an early fusion fashion to enrich diagnostic and anatomical (localization) information. For each modality, we first sample \textit{k}=9 consecutive slices and use them to create a single image, rearranging the selected slices in a $\sqrt{k}\times\sqrt{k}$ grid. $k$ can be set differently depending on the memory availability and $z$-direction resolution of the MRI scan. In our experiments, we optimize this number to appreciate full anatomical information of the pancreas. The two images (one for each modality) are then concatenated along the channel dimension: the resulting tensor, of size $\sqrt{k}H \times \sqrt{k}W \times 2$, with $H\times W$ being the original size of each slice, is provided as input to the transformer. 

Without loss of generality, let us assume that $H=W$. As in~\cite{dosovitskiy2020image}, the input image is then divided into $N$ patches of size $P\times P$, where $N = \frac{kHW}{P^2}$. 
As a result of this procedure, an input image $\mathbf{X} \in \mathbb{R}^{H\times W\times C}$ becomes a sequence of 2D patches $\mathbf{X}_p$ $\in$ $\mathbb{R}^{N\times(P\times P \times C)}$ with $C$ being the channel dimension.
The 2D patches are then flattened into vectors of size $P^2C$ and projected to an \emph{embedding space} of size $D$, obtaining a sequence of \emph{token embeddings}. As a last pre-processing step, learnable positional encodings are summed to token embeddings, producing the actual input data sequence to the transformer. 

We extend the token sequence with a special \emph{class} token, whose state at the output of the transformer describes the overall input image representation for classification purposes~\cite{dosovitskiy2020image,touvron2020training,devlin2018bert}.

Formally, the input $\mathbf{z}_0$ to the transformer is defined as:
\begin{equation}
\mathbf{z}_0 = \left[ \textbf{x}_{\texttt{class}}, \mathbf{x}_p^1 \mathbf{E}, \dots ,   \mathbf{x}_p^N \mathbf{E} \right] + \mathbf{E}_\text{pos} ,
\end{equation}
where each $\mathbf{x}_p^i \in \mathbb{R}^{P^2C}$ is a flattened patch vector, $\mathbf{E} \in$ $\mathbb{R}^{(P^2C) \times D}$ is the embedding matrix and $\mathbf{E}_\text{pos} \in \mathbb{R}^{(N+1) \times D}$ is the positional encoding matrix.

The transformer encoder~\cite{vaswani2017attention} alternates multi-head self-attention and MLP (multilayer perceptron) blocks. These blocks are then intertwined with layer normalization and residual connections (see Fig.~\ref{fig:net}), as follows:
\begin{equation}
\mathbf{z}'_l = \text{MSA}(\text{LN}(\mathbf{z}_{l-1})) +  \mathbf{z}_{l-1} ,
\end{equation}
\begin{equation}
\mathbf{z}_l = \text{MLP}(\text{LN}(\mathbf{z}'_l)) +  \mathbf{z}'_l , \\
\end{equation}
where $l = 1 \cdots L$ identifies the transformer layer, $\text{LN}(\cdot)$ performs layer normalization, $\text{MLP}$ represents a multilayer perceptor, and $\text{MSA}(\cdot)$ computes the standard \emph{query-key-value} multi-head self-attention~\cite{vaswani2017attention}.

At the last transformer layer, the output embedding corresponding to \emph{class} token is finally used for classification into 3 classes, since the MRI dataset includes normal scans, low-grade and high-grade IPMN lesions:
\begin{equation}
\mathbf{y} = \text{LN}(\mathbf{z}^0_{L}) ,
\end{equation}
with $\mathbf{y}$ being the vector of output class scores.

\section{EXPERIMENTAL RESULTS}
\label{sec:exp}

%----------------------------------------------------------------------------------------

\subsection{Dataset}

We evaluate the accuracy of our proposed IPMN risk assessment method in MRI (with both T1 and T2 modalities). We use a total of 139 scans from distinct patients, retrospectively collected at Mayo Clinic~\cite{lalonde2019inn}. Patients have either IPMN cysts detected in their pancreases or they are normal control cases selected to match the IPMN patients. Out of 139 cases, 58 (42\%) were male; mean (standard deviation) age was 65.3 (11.9) years. 22\% had normal pancreas; 34\%, low-grade dysplasia; 14\%, high-grade dysplasia; and 29\%, adenocarcinoma~\cite{corral2019deep}. Two expert radiologists graded the cases in a pathology report after surgery: 0) normal, 1) low-grade IPMN, and 2) high-grade IPMN. We did not consider invasive carcinoma in our analysis as they are outside the scope of IPMN risk stratification.

MRI images were resized (in the transverse plane) to 256$\times$256 pixels. Voxel spacing of MRI scans were varying from 0.468 mm to 1.406 mm. %To minimize uncertainties in MRI scans,
We applied a set of pre-processing steps: N4 bias field correction followed by an edge-preserving Gaussian smoothing, and intensity standardization procedure to normalize MRI scans across patients, scanners, and time. All MRIs were performed using Siemens scanners 1.5 or 3 T (Siemens, Berlin, Germany). 

The experimental procedures involving human subjects described in this paper were approved by the Institutional Review Board.

\subsection{Training Procedure}

We use the Vision-Transformer pre-trained on 300 million images~\cite{sun2017revisiting} and released in~\cite{dosovitskiy2020image}.
During training, we fine-tune all transformers layers with the training data from the MRI dataset. MRI slices are cropped around the pancreas areas by expert physicians for all scans, and each set of 9 consecutive slices, extracted in a sliding window fashion, is arranged in a $3\times3$ grid (from top-left to bottom-right), where each cell of the grid is filled by a 64$\times$64 MRI slice (see Fig.~\ref{fig:net}). Input MRI scans are re-oriented using the RAS axes convention and normalized, individually, between 0 and 1. Data augmentation is performed through random horizontal flipping and random 90-degrees rotation (identically applied to all slices within a grid). We minimize the cross-entropy loss with gradient descent using the Adam optimizer (learning rate: 0.003) and batch size of 8, for a total of 3000 epochs. 
At inference time, we classify each input MRI by feeding the sequence of 9 central slices to the model.

We employ the same training and evaluation procedure for CNN models used as baselines, i.e., DenseNet-121~\cite{huang2017densely}, AlexNet~\cite{krizhevsky2012imagenet} \changeF{ResNet18~\cite{he2016deep}, EfficientNet\_b5~\cite{tan2019efficientnet}  and MobileNet\_v2~\cite{sandler2018mobilenetv2}}.
Experiments are performed on \changeF{a NVIDIA RTX 3090 GPU.}%two NVIDIA Titan X (Pascal architecture) GPUs. 
 The proposed approach was implemented in PyTorch and MONAI; all code will be publicly released upon publication.

\begin{table}[htb!]
    \centering
    \small\addtolength{\tabcolsep}{-2pt}
    \caption{Performance of tested models with 10-fold nested cross-validation. We report results in term of mean $\pm$ standard deviation of metrics computed over all validation folds.}
    \label{tab:results}
    \begin{tabular}{rccc}
    \toprule
    \textbf{Method} & \multicolumn{1}{c}{\textbf{Accuracy}} & \multicolumn{1}{c}{\textbf{Precision}}& \multicolumn{1}{c}{\textbf{Recall}}\\
    %& $\mu$ & $\sigma$ & $\mu$ & $\sigma$ & $\mu$ & $\sigma$\\
    \toprule
    AlexNet  & 0.42 $\pm$ 0.17 &0.37 $\pm$ 0.15 & 0.39 $\pm$ 0.11\\
    DenseNet & 0.51 $\pm$ 0.12 & 0.54 $\pm$ 0.14 & 0.50 $\pm$ 0.14\\
    ResNet18 & 0.53 $\pm$ 0.11 & 0.55 $\pm$ 0.23 & 0.32 $\pm$ 0.08\\
    MobileNet\_v2 & 0.43 $\pm$ 0.11 & 0.54 $\pm$ 0.26 & 0.35 $\pm$ 0.11 \\
    EfficientNet\_b5 & 0.55 $\pm$ 0.10 & 0.60 $\pm$ 0.14 & 0.36 $\pm$ 0.08\\
    \midrule
    \textbf{Ours} & \textbf{0.70} $\pm$ 0.11 & \textbf{0.67} $\pm$ 0.19 & \textbf{0.64} $\pm$ 0.12\\
    \bottomrule
    \end{tabular}
\end{table}

\subsection{Performance}
We perform 10-fold nested cross-validation in order to estimate the accuracy of the proposed approach and the methods under comparison. Results are reported in Table~\ref{tab:results}, where the proposed model largely outperforms the CNN models, confirming the better generalization capabilities of transformer-based architectures compared to standard convolutional models.

\begin{table}[htb!]
    
    \centering
    \small\addtolength{\tabcolsep}{-2pt}
    \caption{Performance of our model using different input data modality with 10-fold nested cross-validation. We report results in terms of mean $\pm$ standard deviation of metrics computed over all validation folds.}
    \label{tab:ablation}
    \begin{tabular}{lccc}
    \toprule
    \textbf{Method} & \multicolumn{1}{c}{\textbf{Accuracy}} & \multicolumn{1}{c}{\textbf{Precision}} & \multicolumn{1}{c}{\textbf{Recall}}\\
   % & $\mu$ & $\sigma$ & $\mu$ & $\sigma$ & $\mu$ & $\sigma$\\
    \toprule
    T1 &  0.53 $\pm$ 0.08 & 0.60 $\pm$ 0.11 & 0.58 $\pm$ 0.14\\
    T2 &  0.64 $\pm$ 0.12 & 0.64 $\pm$ 0.13 & 0.63 $\pm$ 0.11\\
    \midrule
    \multicolumn{4}{c}{\textbf{T1+T2 modalities}}\\
    \midrule
    Late fusion & 0.60  $\pm$0.16 & 0.61 $\pm$ 0.13 & 0.59 $\pm$ 0.11\\
    Early fusion  & \textbf{0.70} $\pm$ 0.11 &\textbf{0.67} $\pm$ 0.19 & \textbf{0.64} $\pm$ 0.12\\
    \bottomrule
    \end{tabular}
\end{table}

We also evaluate the role of early fusion and of combining the T1 and T2 modalities, by assessing classification performance when the model receives only one modality at a time (either T1 or T2) and when performing late fusion. In this case, we train two transformer models, one for each modality, and we then concatenate the two class tokens before classification. Performance is reported in Table~\ref{tab:ablation} that demonstrates how using T1 and T2 in an early fusion setting yields the highest performance. 

\begin{figure*}[ht]
\vspace{.8cm}
    \centering
    \begin{subfigure}{0.15\textwidth}
        \centering
        \includegraphics[width=0.97\textwidth]{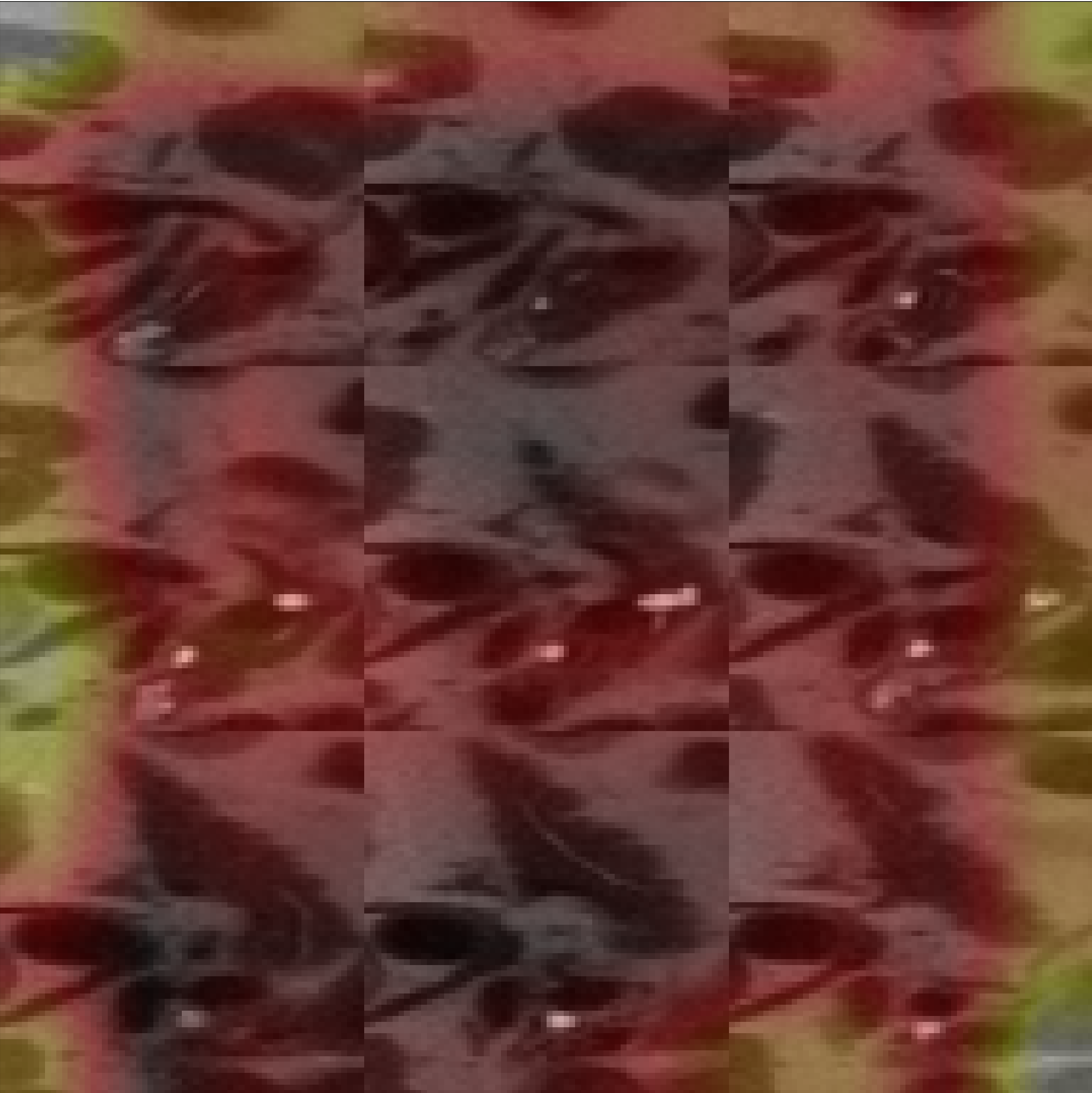}%
    \end{subfigure}%
    \begin{subfigure}{0.15\textwidth}
        \centering
        \includegraphics[width=0.97\textwidth]{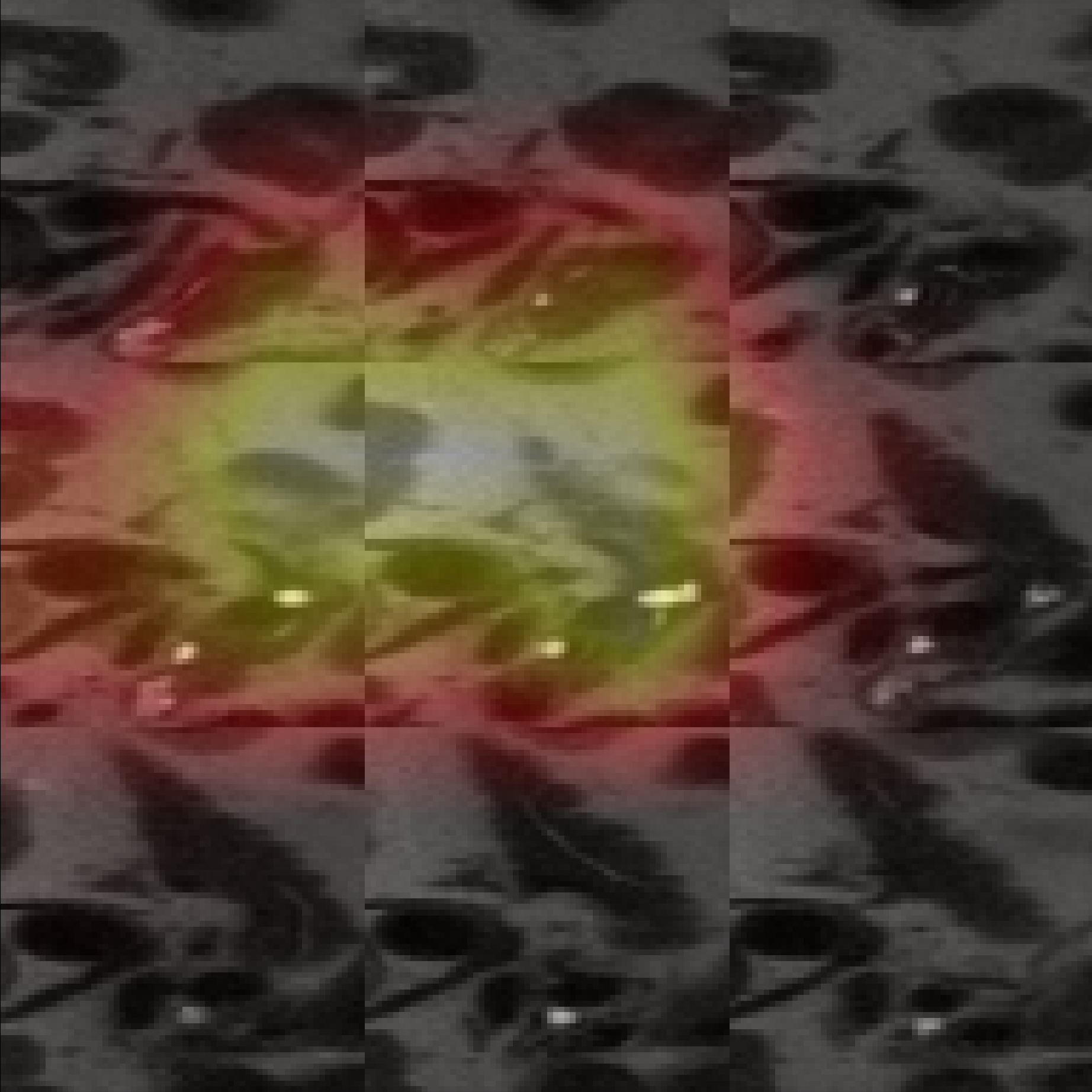}%
    \end{subfigure}%
    \begin{subfigure}{0.15\textwidth}
        \centering
        \includegraphics[width=0.97\textwidth]{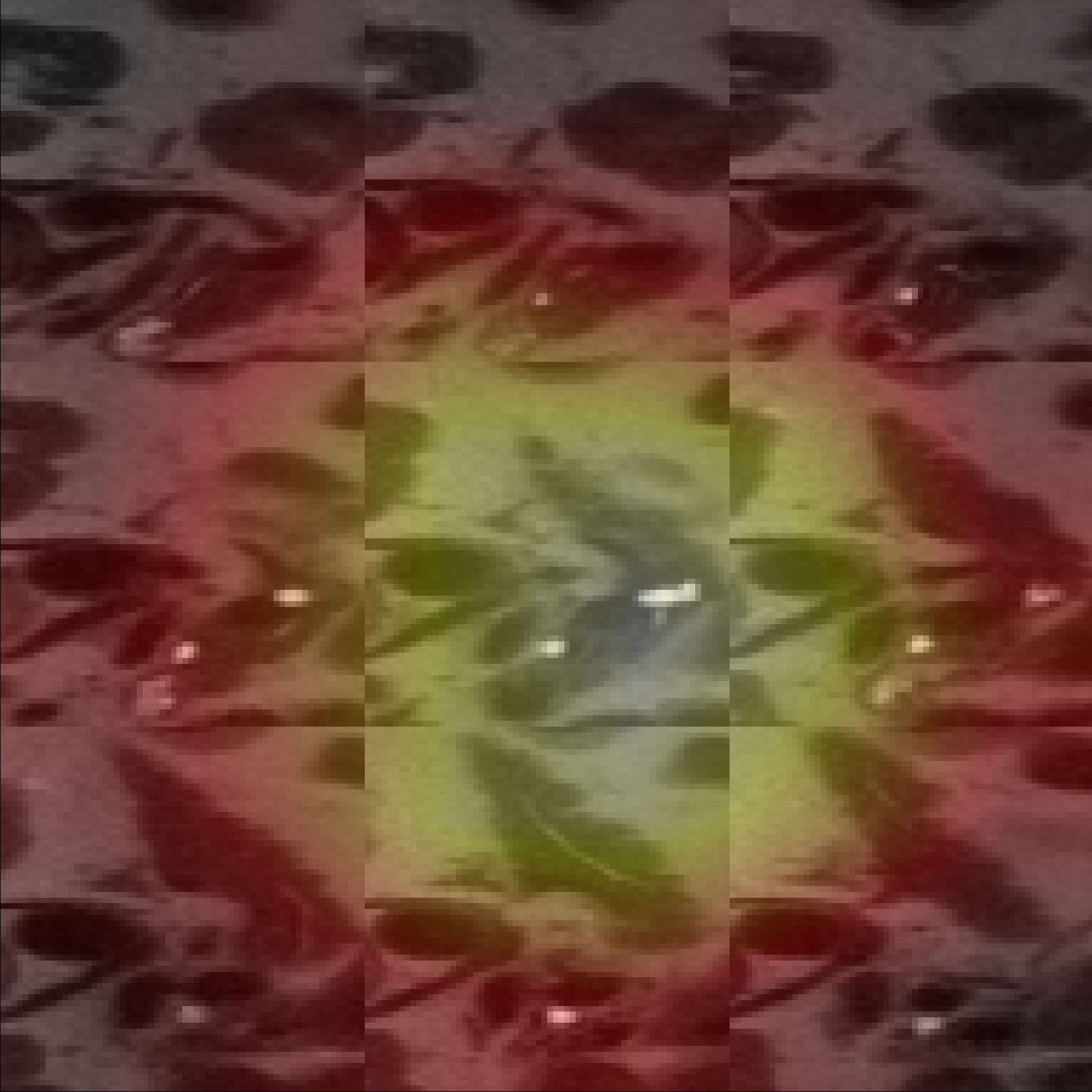}%
    \end{subfigure}%
    \begin{subfigure}{0.15\textwidth}
        \centering
        \includegraphics[width=0.97\textwidth]{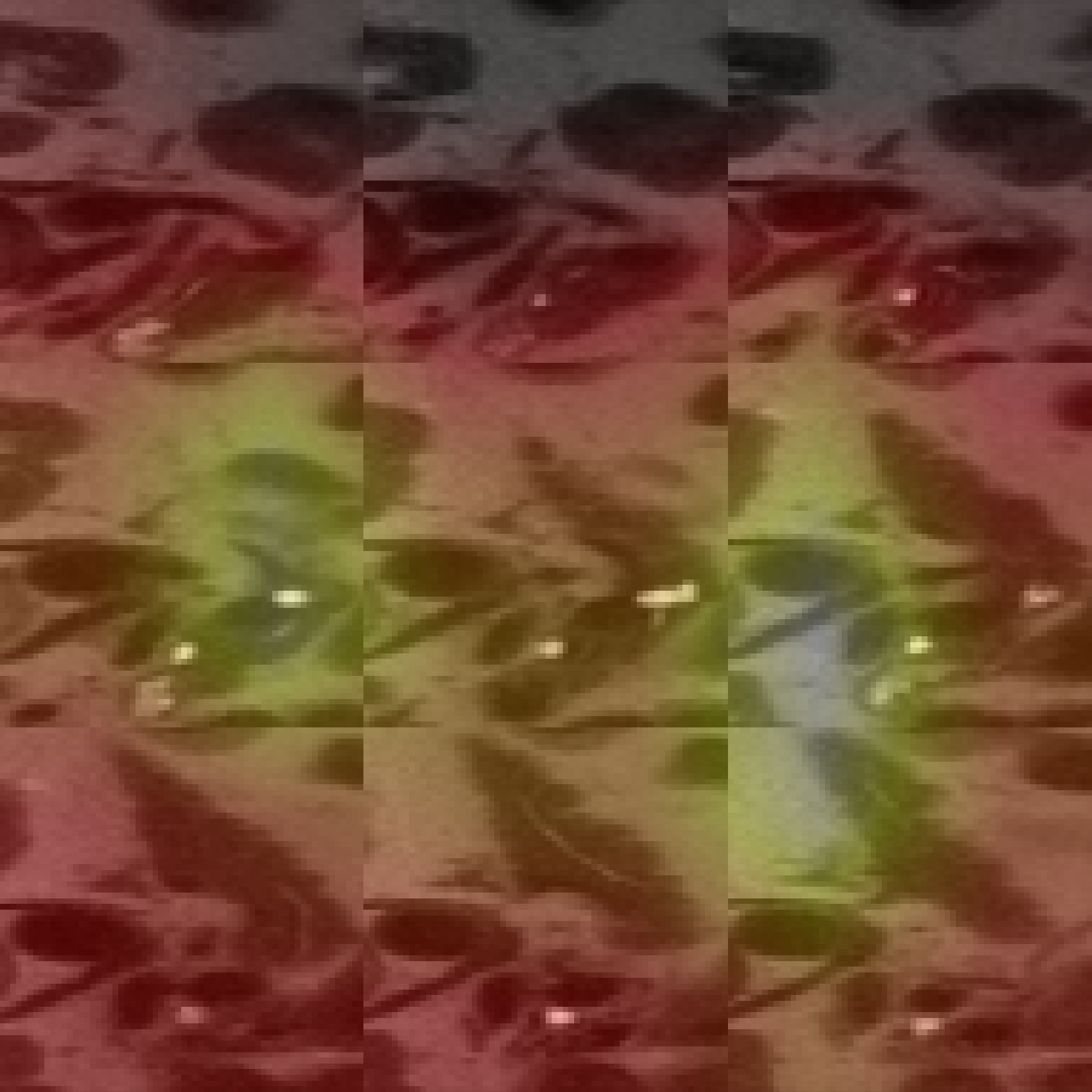}%
    \end{subfigure}%
    \begin{subfigure}{0.15\textwidth}
        \centering
        \includegraphics[width=0.97\textwidth]{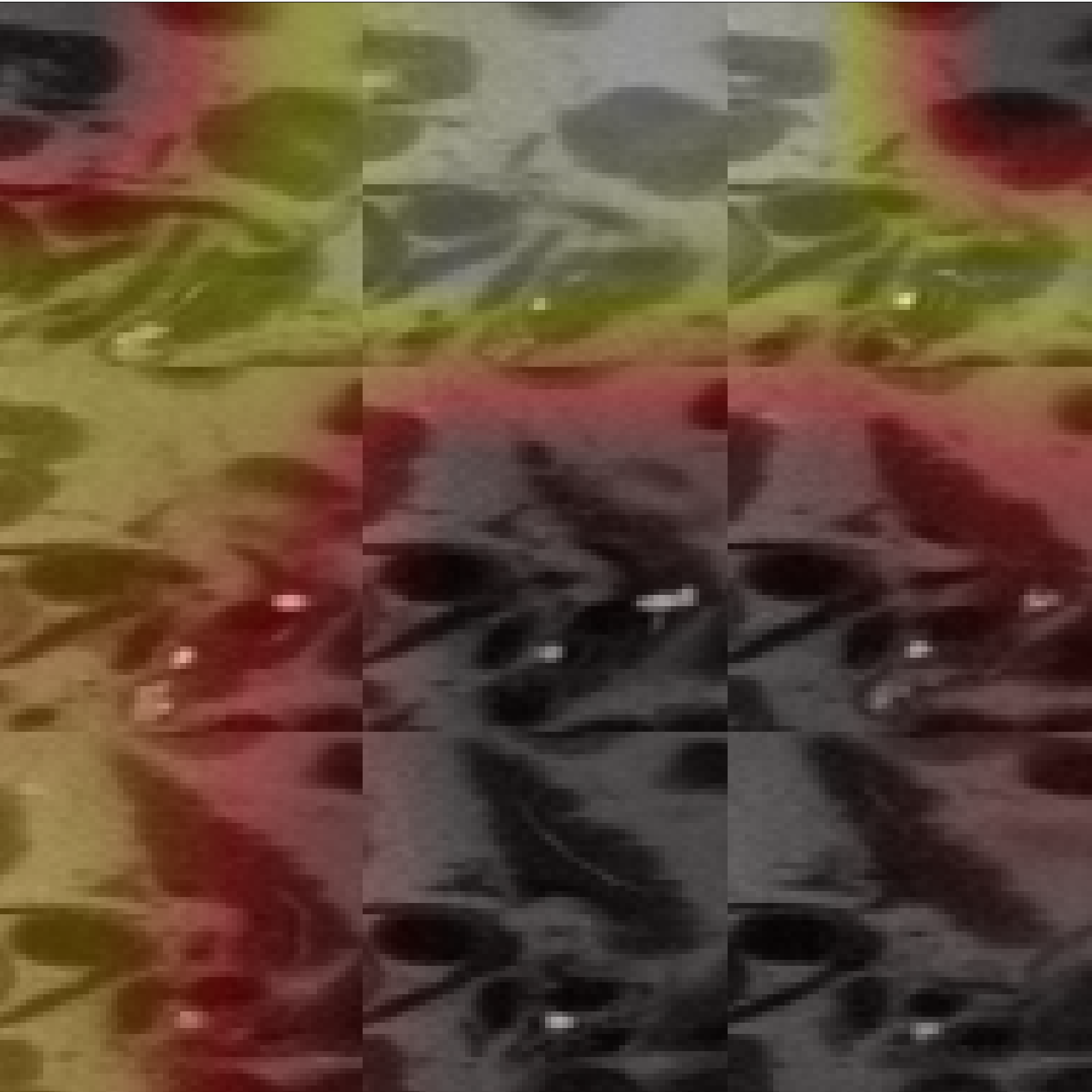}%
    \end{subfigure}%
    \begin{subfigure}{0.15\textwidth}
        \centering
        \includegraphics[width=0.97\textwidth]{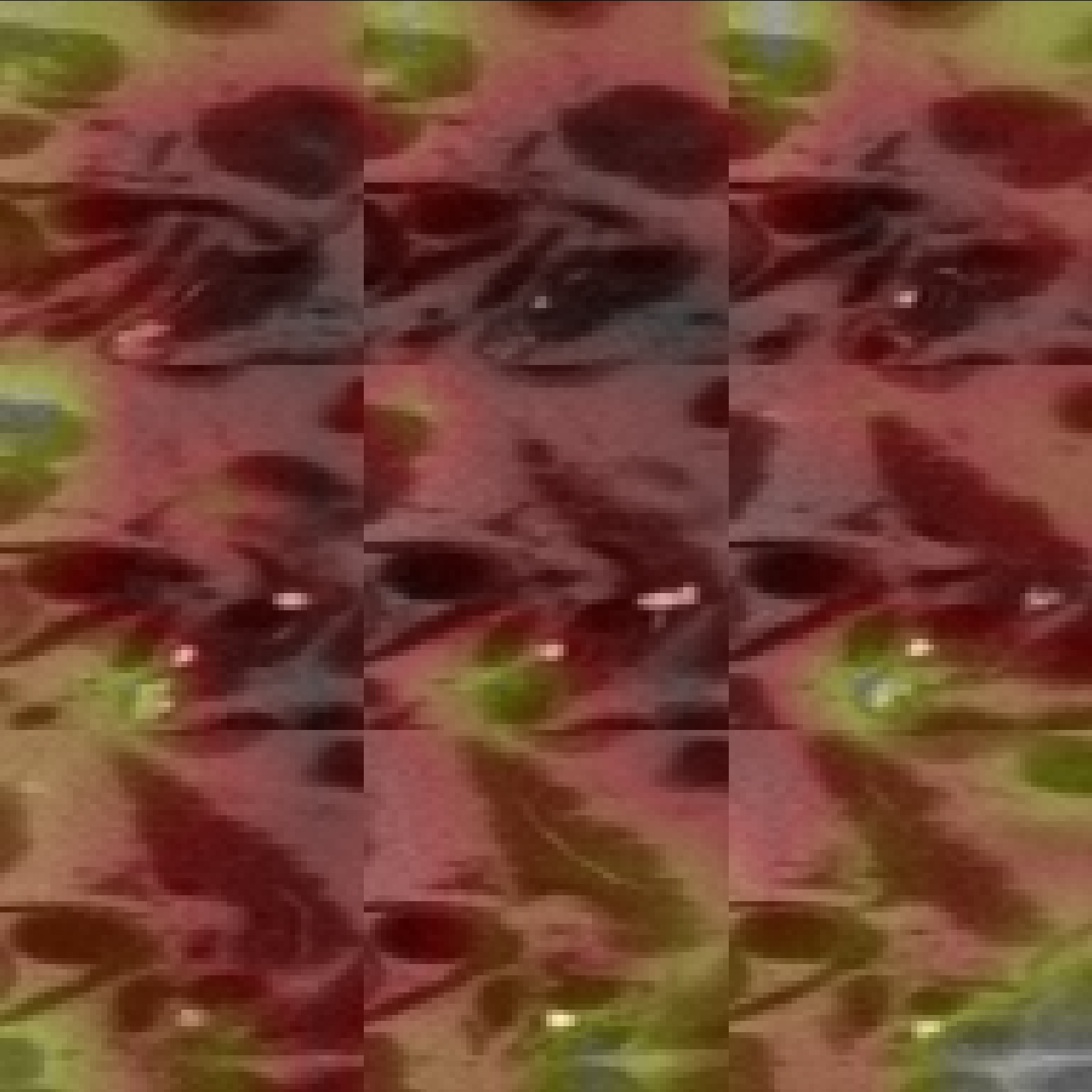}%
    \end{subfigure}%
    \\\vspace{0.1cm}
    \begin{subfigure}{0.15\textwidth}
        \centering
        \includegraphics[width=0.97\textwidth]{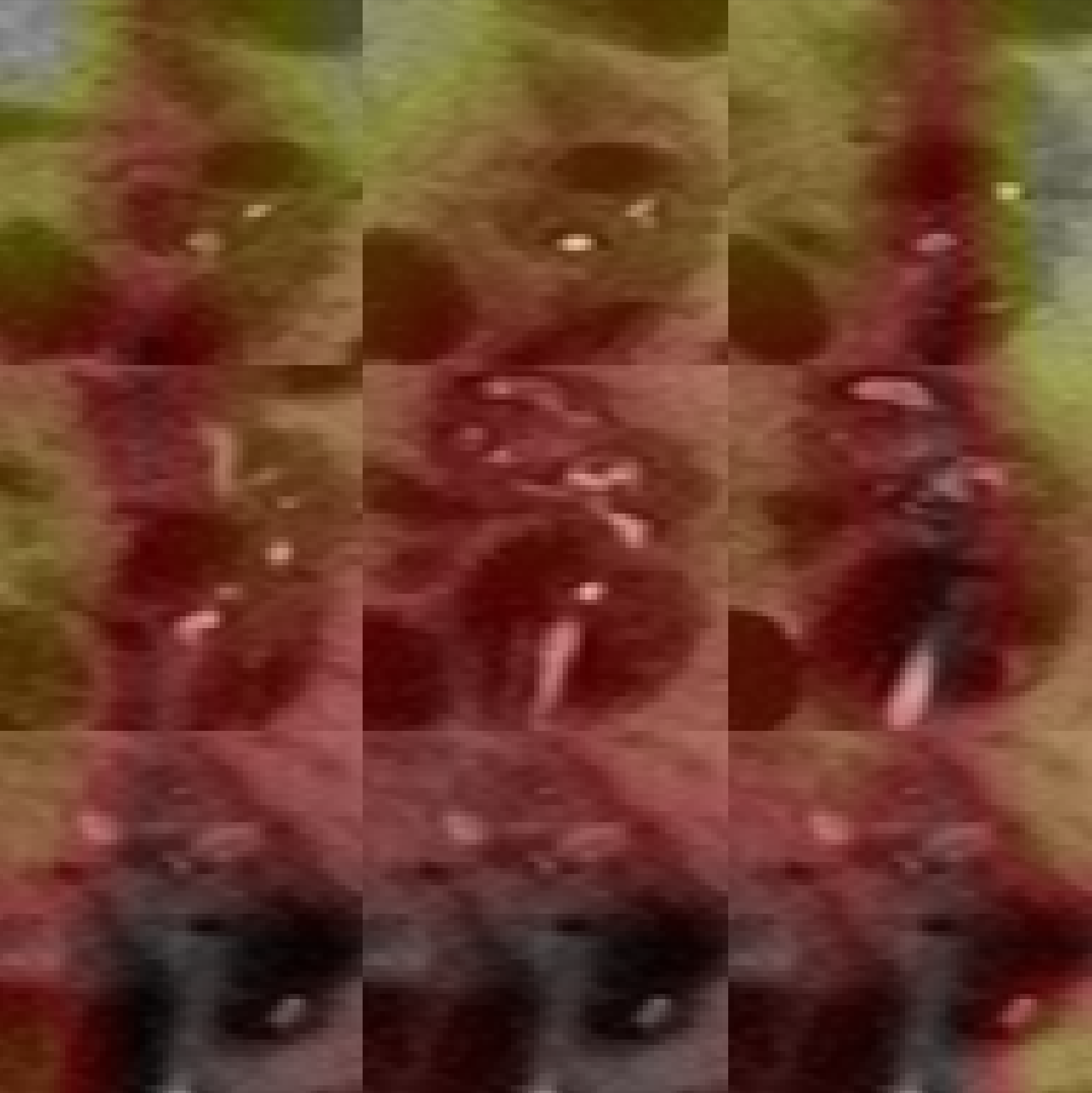}%
        \caption{AlexNet}
    \end{subfigure}%
        \begin{subfigure}{0.15\textwidth}
        \centering
        \includegraphics[width=0.97\textwidth]{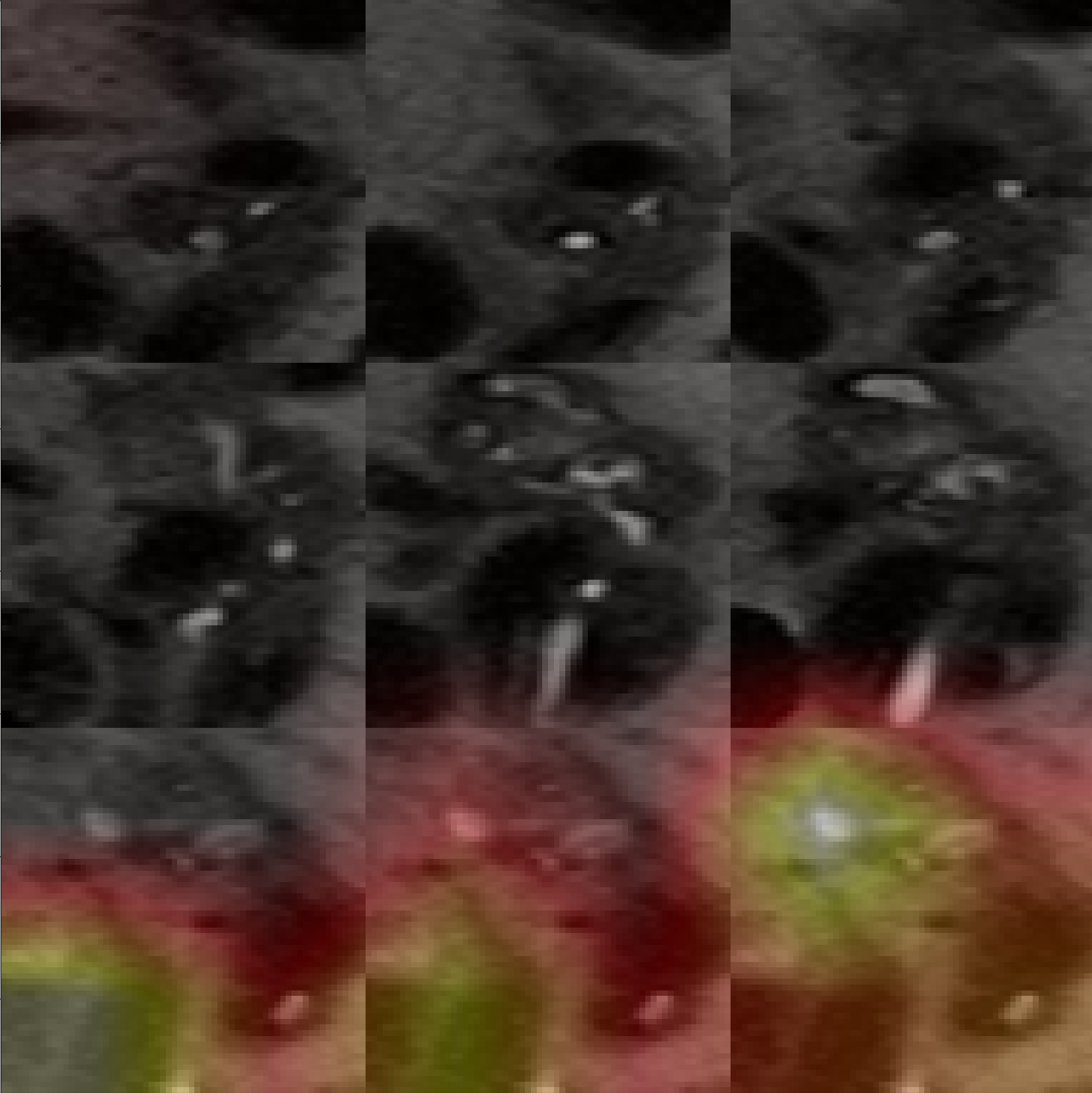}%
        \caption{DenseNet}
    \end{subfigure}%
    \begin{subfigure}{0.15\textwidth}
        \centering
        \includegraphics[width=0.97\textwidth]{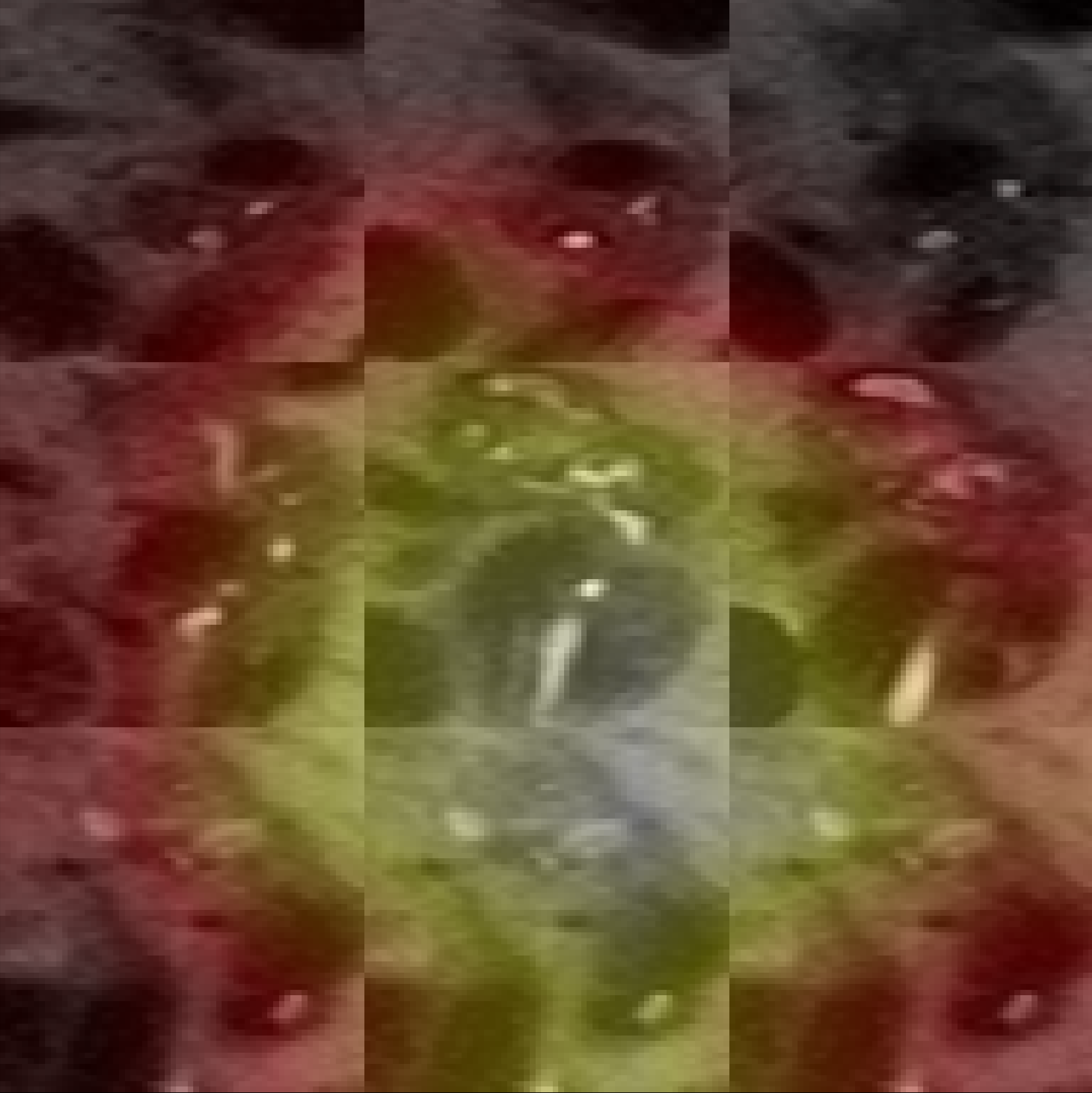}%
        \caption{MobileNet\_v2}
    \end{subfigure}%
    \begin{subfigure}{0.15\textwidth}
        \centering
        \includegraphics[width=0.97\textwidth]{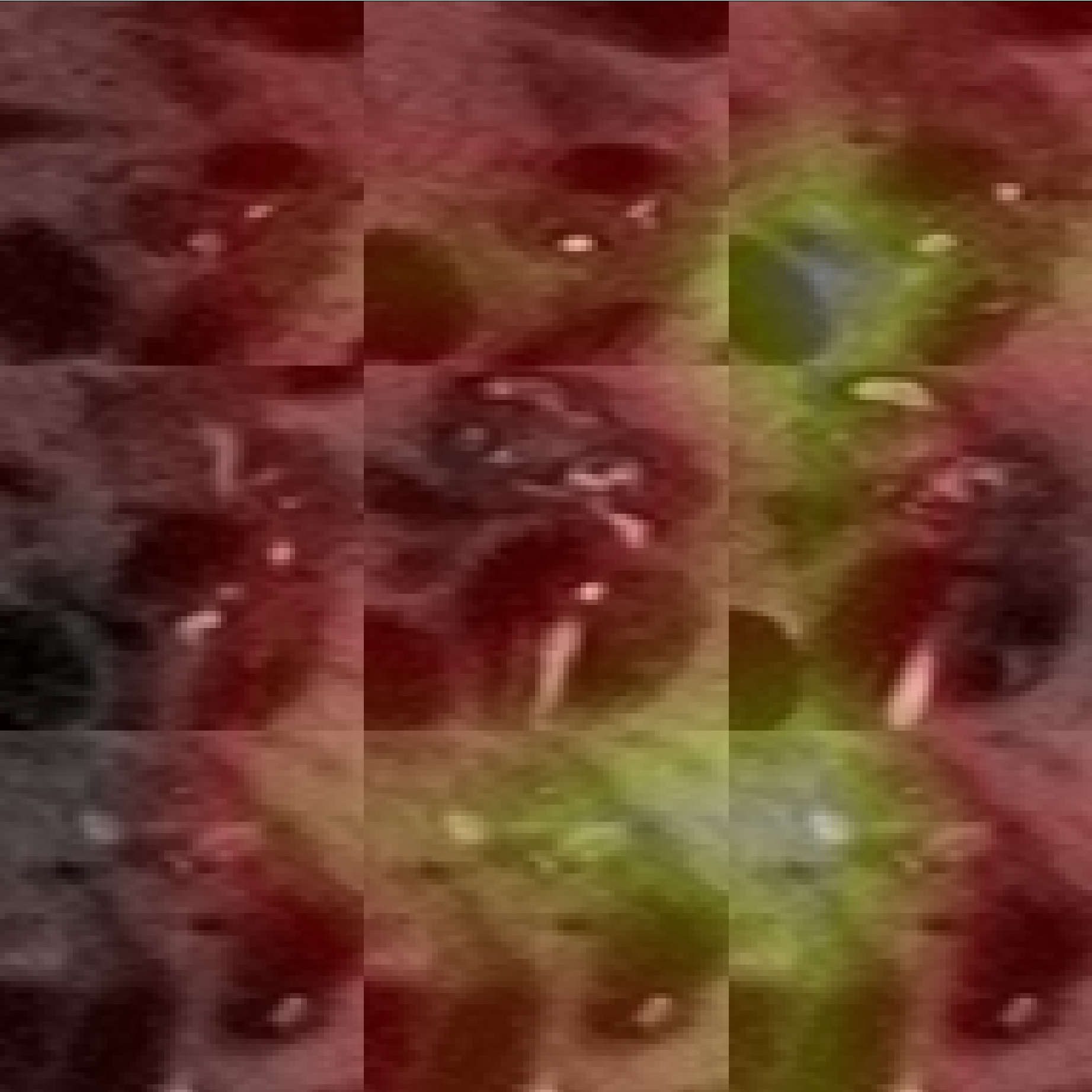}%
        \caption{ResNet18}
    \end{subfigure}%
    \begin{subfigure}{0.15\textwidth}
        \centering
        \includegraphics[width=0.97\textwidth]{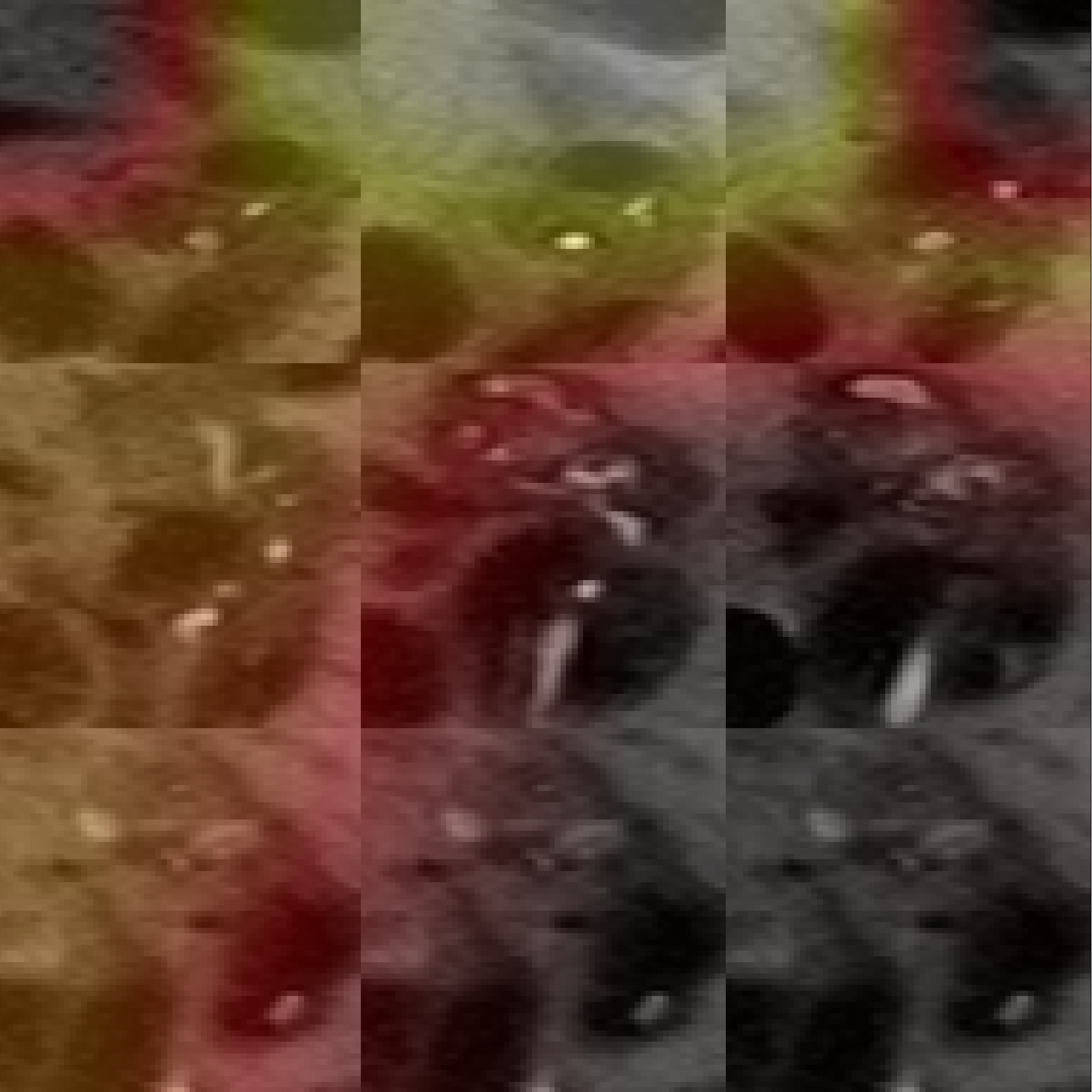}%
        \caption{EfficientNet\_b5}
    \end{subfigure}%
    \begin{subfigure}{0.15\textwidth}
        \centering
        \includegraphics[width=0.97\textwidth]{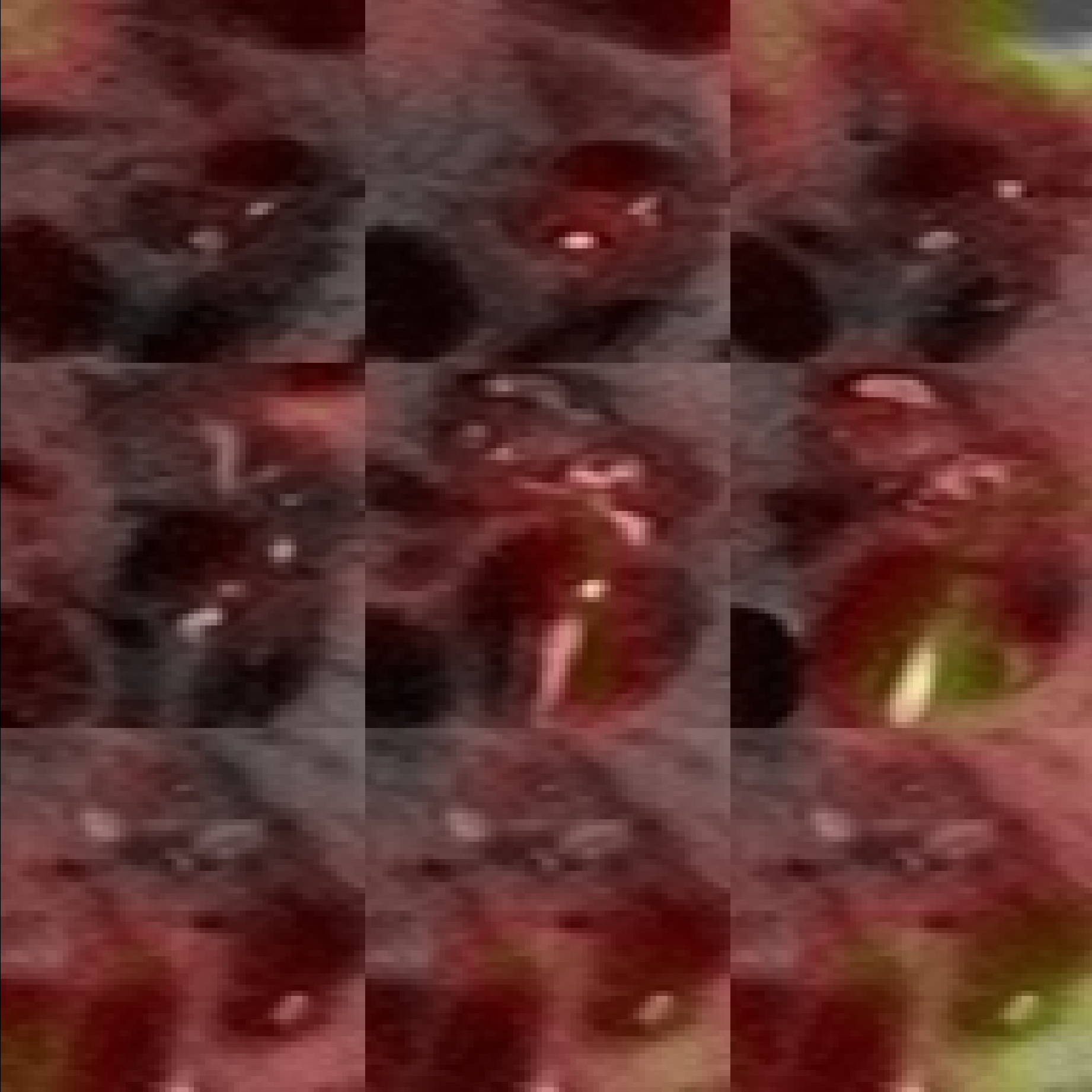}%
        \caption{Our method}
    \end{subfigure}%
    \caption{Comparison between the attention maps of AlexNet, DenseNet-121 and our model in case of correct (top row) and erroneous (bottom row) predictions  on a 3$\times$3 grid of MRI images.}
    \label{fig:interpret2}
\end{figure*}

It has to be noted that, comparing on the same dataset, the performance achieved by our transformer-based approach is slightly lower than those obtained in~\cite{lalonde2019inn}, i.e., about 73\%. 
However, the architecture in~\cite{lalonde2019inn} was specifically designed and tuned for solving the IPMN classification problem, while our transformer architecture is general, designed for natural image classification and applied directly without significant architectural changes to IPMN classification problem. This is remarkable, as we demonstrate that a general architecture performs similarly to an ad-hoc one for a complex task with limited training data.

\subsection{Interpretability of results}
\label{ssec:interpr}
\begin{figure}[hb!]
\centering
    \begin{subfigure}{0.32\linewidth}
        \includegraphics[width=0.98\linewidth]{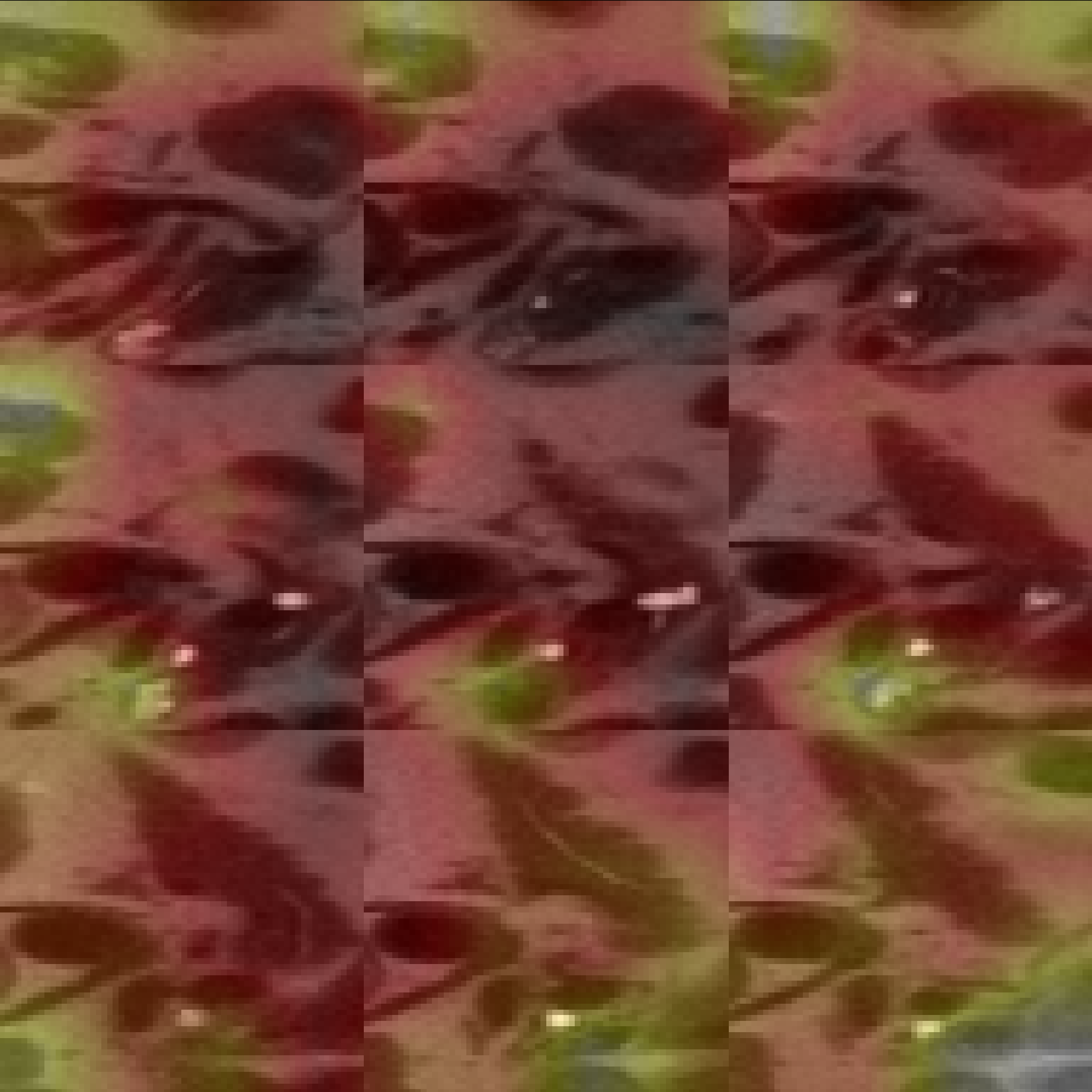}
    \end{subfigure}
    \begin{subfigure}{0.32\linewidth}
        \includegraphics[width=0.98\linewidth]{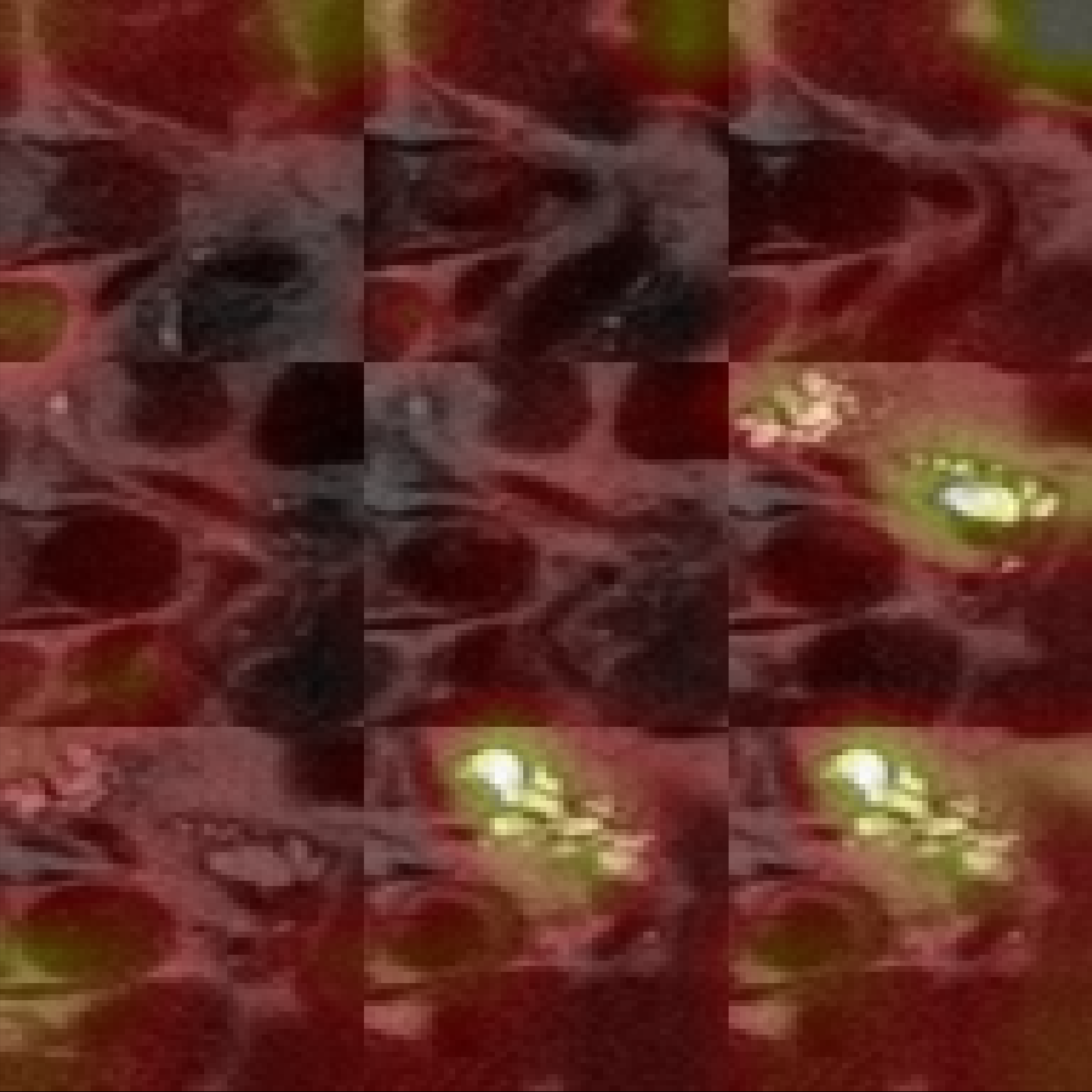}
    \end{subfigure}
    \begin{subfigure}{0.32\linewidth}
        \includegraphics[width=0.98
        \linewidth]{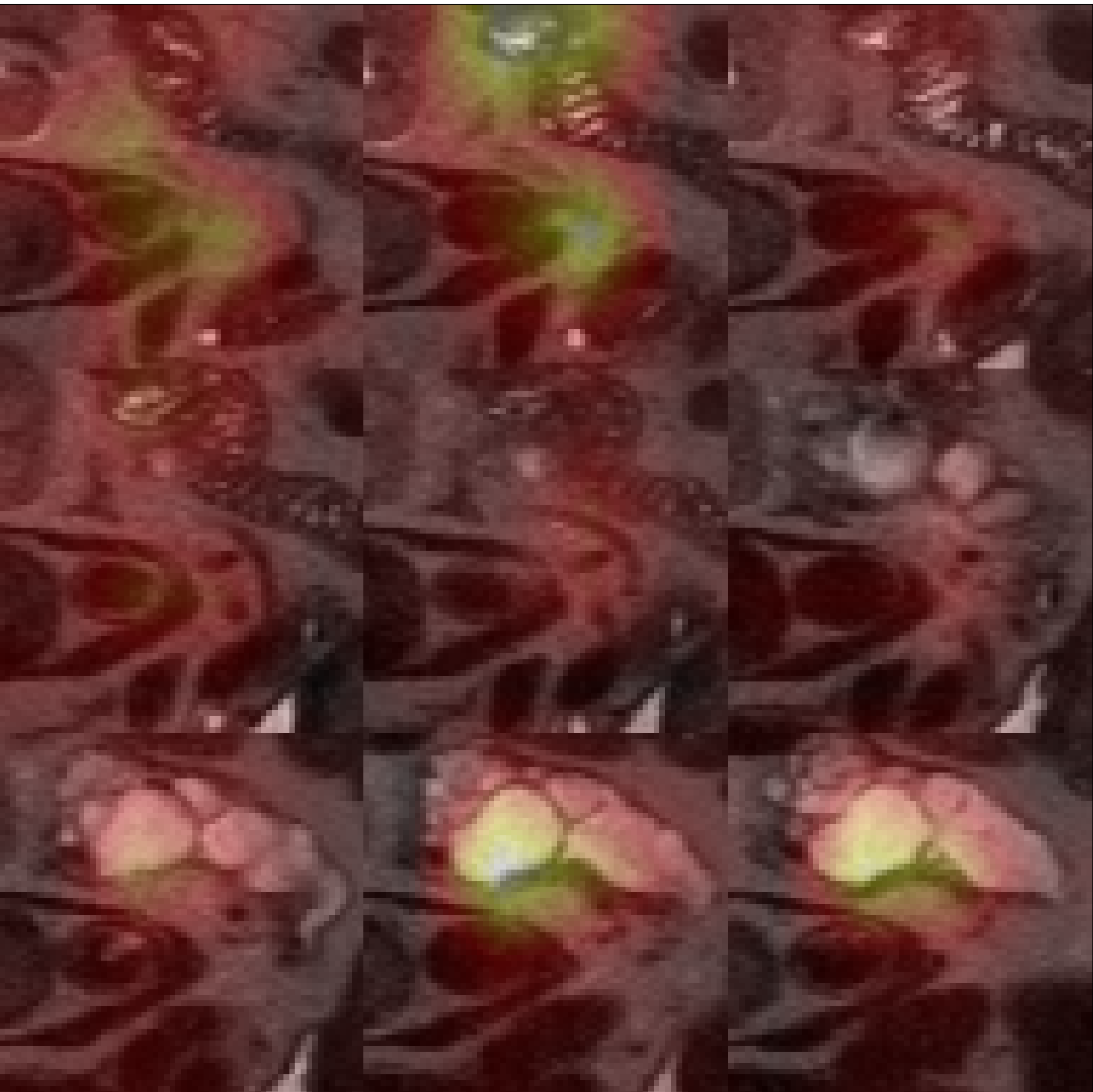}
    \end{subfigure}
    \caption{Attention maps of our transformer-based classifier on 3$\times$3 grid of MRI images for correct IPMN classification.}
    \label{fig:interpret}
\end{figure}

Transformers allow for a more direct interpretation of its internal representations through visualizing the attention weights~\cite{dosovitskiy2020image}, thus supporting the sought interpretability necessary in safety-critical contexts as the medical domain one. 
We apply Attention Rollout~\cite{abnar2020quantifying} to track down the information propagated from the input layer to the embeddings in the higher layers. Thus, we average attention weights of all heads of each transformer layer and then multiply these averages across all layers. Fig.~\ref{fig:interpret} shows some examples of interpretabilty maps in cases of correct cyst classification. 

It can be observed how our transformer-based model focuses its attention mainly on cysts, thus it provides robust predictions.
Conversely, CNN-based models lead to weak decisions, as their attention maps (estimated using GradCam~\cite{selvaraju2017grad}) reveal that features not strictly related to cysts are used for classification (see Fig.~\ref{fig:interpret2}, top row). 
Finally, although our model fails in some cases, as demonstrated by the classification accuracy in Tab.~\ref{tab:results}, its attention maps often point to the correct cyst regions (see Fig.~\ref{fig:interpret2}, bottom row); thus, the wrong prediction is due to either using directly raw data, rather than a more powerful representation, or lack of enough training data.

\section{Conclusion}
\label{sec:conclusion}

In this work, our overall goal was to classify pancreas (IPMN) cysts automatically. We utilized \textit{transformers} for the first time for pancreas risk predictions and obtained promising results that can be used for MRI-based IPMN risk stratification routinely. Compared to the (few) existing methods, transformers showed higher performance overall. We found that training transformer for IPMN risk stratification is easier than conventional CNN based systems and generalizes better.
Furthermore, the proposed transformer-based classifier allows for better interpretation of results than standard CNNs, revealing how it employs cues exclusively related to cysts, providing more robustness to the automated diagnosis than the comparing methods.
These findings highlight the contribution that transformers can give to the future research in medical image understanding, in general, and IPMN classification, in particular, beside contributing the recent AI research efforts towards universal architectures.

\addtolength{\textheight}{0cm}   % This command serves to balance the column lengths

%%%%%%%%%%%%%%%%%%%%%%%%%%%%%%%%%%%%%%%%%%%%%%%%%%%%%%%%%%%%%%%%%%%%%%%%%%%%%%%%

%%%%%%%%%%%%%%%%%%%%%%%%%%%%%%%%%%%%%%%%%%%%%%%%%%%%%%%%%%%%%%%%%%%%%%%%%%%%%%%%
%\section*{APPENDIX}


\section*{ACKNOWLEDGMENT}
This research was partially supported by the following grants: 1) Rehastart funded by PO FESR 2014-2020 Sicilia, Azione 1.1.5; project number 08ME6201000222, CUP: G79J18000610007); 2) the “Adaptive Brain-Derived Artificial Intelligence Methods for Event Detection” – BrAIn funded by the “Programma per la ricerca di ateneo UNICT 2020–22 linea 3”; 3) NIH grants R01-CA246704-01 and R01-CA240639-01. 

%This project is partially supported by REHASTART and iHosp %projects.

%The preferred spelling of the word ÒacknowledgmentÓ in America is without an ÒeÓ after the ÒgÓ. Avoid the stilted expression, ÒOne of us (R. B. G.) thanks . . .Ó  Instead, try ÒR. B. G. thanksÓ. Put sponsor acknowledgments in the unnumbered footnote on the first page.

%%%%%%%%%%%%%%%%%%%%%%%%%%%%%%%%%%%%%%%%%%%%%%%%%%%%%%%%%%%%%%%%%%%%%%%%%%%%%%%%

%References are important to the reader; therefore, each citation must be complete and correct. If at all possible, references should be commonly available publications.


\newpage
\begin{thebibliography}{99}

\bibitem{society2021cancer}
American~Cancer Society,
\newblock ``Cancer facts \& figures,''
\newblock {\em American Cancer Society}, 2021.

\bibitem{vaswani2017attention}
Ashish Vaswani, Noam Shazeer, Niki Parmar, Jakob Uszkoreit, Llion Jones,
  Aidan~N Gomez, Lukasz Kaiser, and Illia Polosukhin,
\newblock ``Attention is all you need,''
\newblock {\em arXiv preprint arXiv:1706.03762}, 2017.

\bibitem{dosovitskiy2020image}
Alexey Dosovitskiy, Lucas Beyer, Alexander Kolesnikov, Dirk Weissenborn,
  Xiaohua Zhai, Thomas Unterthiner, Mostafa Dehghani, Matthias Minderer, Georg
  Heigold, Sylvain Gelly, et~al.,
\newblock ``An image is worth 16x16 words: Transformers for image recognition
  at scale,''
\newblock {\em arXiv preprint arXiv:2010.11929}, 2020.

\bibitem{lalonde2019inn}
Rodney LaLonde, Irene Tanner, Katerina Nikiforaki, Georgios~Z Papadakis, Pujan
  Kandel, Candice~W Bolan, Michael~B Wallace, and Ulas Bagci,
\newblock ``Inn: inflated neural networks for ipmn diagnosis,''
\newblock in {\em International Conference on Medical Image Computing and
  Computer-Assisted Intervention}. Springer, 2019, pp. 101--109.

\bibitem{zhou2017deep}
Yuyin Zhou, Lingxi Xie, Elliot~K Fishman, and Alan~L Yuille,
\newblock ``Deep supervision for pancreatic cyst segmentation in abdominal ct
  scans,''
\newblock in {\em International conference on medical image computing and
  computer-assisted intervention}. Springer, 2017, pp. 222--230.

\bibitem{kuwahara2019usefulness}
Takamichi Kuwahara, Kazuo Hara, Nobumasa Mizuno, Nozomi Okuno, Shimpei
  Matsumoto, Masahiro Obata, Yusuke Kurita, Hiroki Koda, Kazuhiro Toriyama,
  Sachiyo Onishi, et~al.,
\newblock ``Usefulness of deep learning analysis for the diagnosis of
  malignancy in intraductal papillary mucinous neoplasms of the pancreas,''
\newblock {\em Clinical and translational gastroenterology}, vol. 10, no. 5,
  2019.

\bibitem{gorris2021artificial}
Myrte Gorris, Sanne~A Hoogenboom, Michael~B Wallace, and Jeanin~E van Hooft,
\newblock ``Artificial intelligence for the management of pancreatic
  diseases,''
\newblock {\em Digestive Endoscopy}, vol. 33, no. 2, pp. 231--241, 2021.

\bibitem{hanania2016quantitative}
Alexander~N Hanania, Leonidas~E Bantis, Ziding Feng, Huamin Wang, Eric~P Tamm,
  Matthew~H Katz, Anirban Maitra, and Eugene~J Koay,
\newblock ``Quantitative imaging to evaluate malignant potential of ipmns,''
\newblock {\em Oncotarget}, vol. 7, no. 52, pp. 85776, 2016.

\bibitem{gazit2017quantification}
Lior Gazit, Jayasree Chakraborty, Marc Attiyeh, Liana Langdon-Embry, Peter~J
  Allen, Richard~KG Do, and Amber~L Simpson,
\newblock ``Quantification of ct images for the classification of high-and
  low-risk pancreatic cysts,''
\newblock in {\em Medical Imaging 2017: Computer-Aided Diagnosis}.
  International Society for Optics and Photonics, 2017, vol. 10134, p. 101340X.

\bibitem{hussein2018deep}
Sarfaraz Hussein, Pujan Kandel, Juan~E Corral, Candice~W Bolan, Michael~B
  Wallace, and Ulas Bagci,
\newblock ``Deep multi-modal classification of intraductal papillary mucinous
  neoplasms (ipmn) with canonical correlation analysis,''
\newblock in {\em 2018 IEEE 15th International Symposium on Biomedical Imaging
  (ISBI 2018)}. IEEE, 2018, pp. 800--804.

\bibitem{corral2019deep}
Juan~E Corral, Sarfaraz Hussein, Pujan Kandel, Candice~W Bolan, Ulas Bagci, and
  Michael~B Wallace,
\newblock ``Deep learning to classify intraductal papillary mucinous neoplasms
  using magnetic resonance imaging,''
\newblock {\em Pancreas}, vol. 48, no. 6, pp. 805--810, 2019.

\bibitem{devlin2018bert}
Jacob Devlin, Ming-Wei Chang, Kenton Lee, and Kristina Toutanova,
\newblock ``Bert: Pre-training of deep bidirectional transformers for language
  understanding,''
\newblock {\em arXiv preprint arXiv:1810.04805}, 2018.

\bibitem{radford}
Alec Radford, Jeffrey Wu, Rewon Child, David Luan, Dario Amodei, and Ilya
  Sutskever,
\newblock ``Language models are unsupervised multitask learners,''
\newblock 2018.

\bibitem{carion2020end}
Nicolas Carion, Francisco Massa, Gabriel Synnaeve, Nicolas Usunier, Alexander
  Kirillov, and Sergey Zagoruyko,
\newblock ``End-to-end object detection with transformers,''
\newblock in {\em European Conference on Computer Vision}. Springer, 2020, pp.
  213--229.

\bibitem{touvron2020training}
Hugo Touvron, Matthieu Cord, Matthijs Douze, Francisco Massa, Alexandre
  Sablayrolles, and Herv{\'e} J{\'e}gou,
\newblock ``Training data-efficient image transformers \& distillation through
  attention,''
\newblock {\em arXiv preprint arXiv:2012.12877}, 2020.

\bibitem{sun2017revisiting}
Chen Sun, Abhinav Shrivastava, Saurabh Singh, and Abhinav Gupta,
\newblock ``Revisiting unreasonable effectiveness of data in deep learning  era,''
\newblock in {\em Proceedings of the IEEE international conference on computer  vision}, 2017, pp. 843--852.

\bibitem{abnar2020quantifying}
Samira Abnar and Willem Zuidema,
\newblock ``Quantifying attention flow in transformers,''
\newblock {\em arXiv preprint arXiv:2005.00928}, 2020.

\bibitem{selvaraju2017grad}
Ramprasaath~R Selvaraju, Michael Cogswell, Abhishek Das, Ramakrishna Vedantam,
  Devi Parikh, and Dhruv Batra,
\newblock ``Grad-cam: Visual explanations from deep networks via gradient-based
  localization,''
\newblock in {\em Proceedings of the IEEE international conference on computer
  vision}, 2017, pp. 618--626.
  
\bibitem{chen2021transunet}
Chen, Jieneng and Lu, Yongyi and Yu, Qihang and Luo, Xiangde and Adeli, Ehsan and Wang, Yan and Lu, Le and Yuille, Alan L and Zhou, Yuyin
\newblock ``Transunet: Transformers make strong encoders for medical image segmentation,''
\newblock {\em arXiv preprint arXiv:2102.04306}, 2021.

\bibitem{hatamizadeh2022unetr}
Hatamizadeh, Ali and Tang, Yucheng and Nath, Vishwesh and Yang, Dong and Myronenko, Andriy and Landman, Bennett and Roth, Holger R and Xu, Daguang
\newblock ``Unetr: Transformers for 3d medical image segmentation,''
\newblock in {\em Proceedings of the IEEE/CVF Winter Conference on Applications of Computer Vision},2022,  pp. 574--584.

\bibitem{xie2021cotr}
Xie, Yutong and Zhang, Jianpeng and Shen, Chunhua and Xia, Yong
\newblock ``CoTr: Efficiently Bridging CNN and Transformer for 3D Medical Image Segmentation,''
\newblock {\em arXiv preprint arXiv:2103.03024}, 2021

\bibitem{huang2017densely}
Huang, Gao and Liu, Zhuang and Van Der Maaten, Laurens and Weinberger, Kilian Q
\newblock ``Densely connected convolutional networks''
\newblock in {\em Proceedings of the IEEE conference on computer vision and pattern recognition},2017, pp. 4700--4708

\bibitem{krizhevsky2012imagenet}
Krizhevsky, Alex and Sutskever, Ilya and Hinton, Geoffrey E
\newblock ``Imagenet classification with deep convolutional neural networks''
\newblock in {\em Advances in neural information processing systems},2012, 25, pp. 1097--1105

\bibitem{he2016deep}
He, Kaiming and Zhang, Xiangyu and Ren, Shaoqing and Sun, Jian
\newblock ``Deep residual learning for image recognition''
\newblock in {\em Proceedings of the IEEE conference on computer vision and pattern recognition},2016, pp. 770--778

\bibitem{tan2019efficientnet}
Tan, Mingxing and Le, Quoc
\newblock ``Efficientnet: Rethinking model scaling for convolutional neural networks''
\newblock in {\em International Conference on Machine Learning}, 2019, pp. 6105--6114 

\bibitem{sandler2018mobilenetv2}
Sandler, Mark and Howard, Andrew and Zhu, Menglong and Zhmoginov, Andrey and Chen, Liang-Chieh
\newblock ``Mobilenetv2: Inverted residuals and linear bottlenecks''
\newblock in {\em Proceedings of the IEEE conference on computer vision and pattern recognition},2018, pp. 4510--4520

\end{thebibliography}
\end{document}